\documentclass{article}

 \usepackage[preprint]{submission}

\usepackage[utf8]{inputenc} % allow utf-8 input
\usepackage[T1]{fontenc}    % use 8-bit T1 fonts
\usepackage{hyperref}       % hyperlinks
\usepackage{url}            % simple URL typesetting
\usepackage{booktabs}       % professional-quality tables
\usepackage{amsfonts}       % blackboard math symbols
\usepackage{nicefrac}       % compact symbols for 1/2, etc.
\usepackage{microtype}      % microtypography
\usepackage{xcolor}         % colors
\usepackage{comment}

\title{Computational Hermeneutics: \\Evaluating Generative AI as a Cultural Technology}

\author{%
  Cody Kommers \\
  The Alan Turing Institute\\
  \texttt{ckommers@turing.ac.uk} \\
   \And
   Ruth Ahnert \\
   Queen Mary\\ University of London \\
   \And
   Maria Antoniak \\
   University of Colorado \\
   \And
   Emmanouil Benetos \\
   Queen Mary\\ University of London \\
   \And
   Steve Benford \\
   University of Nottingham \\
   \And
   Mercedes Bunz \\
   King's College\\ London \\
   \And
   Baptiste Caramiaux \\
   Sorbonne Université
   \And
   Shauna Concannon \\
   Durham University \\
   \And
   Martin Disley \\
   University of Edinburgh \\
   \And
   James Dobson \\
   Dartmouth College\\
   \And
   Yali Du \\
   King's College London
   \And
   Edgar Duéñez-Guzmán \\
   Gibran AI\\
   \And
   Kerry Francksen \\
   University of Coventry \\
   \And
   Evelyn Gius \\
   Technische Universität Darmstadt \\
   \And
   Jonathan Gray \\
   King's College London \\
   \And
   Ryan Heuser \\
   University of Cambridge \\
   \And
   Sarah Immel \\
   University of Edinburgh \\
   \And
   Richard Jean So \\
   McGill University \\
   \And
   Sang Leigh \\
   Cornell University \\
   \And
   Dalaki Livingston \\
   University of Utah \\
   \And
   Hoyt Long \\
   University of Chicago \\
   \And
   Meredith Martin \\
   Princeton University \\
   \And
   Georgia Meyer \\
   London School of Economics \\
   \And
   Daniela Mihai \\
   University of Southampton \\
   \And
   Ashley Noel-Hirst \\
   The Alan Turing Institute \\
   \And
   Kirsten Ostherr \\
   Rice University \\
   \And
   Deven Parker \\
   University of Glasgow \\
   \And
   Yipeng Qin \\
   Cardiff University \\
   \And
   Jessica Ratcliff \\
   Cornell University \\
   \And
   Emily Robinson \\
   University of Exeter \\
   \And
   Karina Rodriguez \\
   University of Brighton \\
   \And
   Adam Sobey \\
   The Alan Turing Institute \& \\
   University of Southampton \\
   \And
   Ted Underwood \\
   University of Illinois\\
   Urbana-Champaign \\
   \And
   Aditya Vashistha \\
   Cornell University \\
   \And
   Matthew Wilkens \\
   Cornell University \\
   \And
   Wu Youyou \\
   University College\\London \\
   \And
   Zheng Yuan \\
   University of Sheffield \\
   \And
   Drew Hemment \\
   The Alan Turing Institute \&\\
   University of Edinburgh \\
}

\begin{document}

\maketitle

\newpage
\begin{abstract}
Generative AI (GenAI) systems are increasingly recognized as cultural technologies, yet current evaluation frameworks often treat culture as a variable to be measured rather than fundamental to the system's operation. Drawing on hermeneutic theory from the humanities, we argue that GenAI systems function as ``context machines'' that must inherently address three interpretive challenges: situatedness (meaning only emerges in context), plurality (multiple valid interpretations coexist), and ambiguity (interpretations naturally conflict). We present computational hermeneutics as an emerging framework offering an interpretive account of what GenAI systems do, and how they might do it better. We offer three principles for hermeneutic evaluation---that benchmarks should be iterative, not one-off; include people, not just machines; and measure cultural context, not just model output. This perspective offers a nascent paradigm for designing and evaluating contemporary AI systems: shifting from standardized questions about accuracy to contextual ones about meaning. 

\end{abstract}

\section{Introduction}

Generative AI (GenAI) systems are cultural and social technologies \cite{farrell2025large,bender2021dangers,sorensen2024position,klein2025provocations}. Every aspect of these systems---from the social data they are trained on, to the benchmarks by which they are evaluated, to the societal effects of their outputs---depends on the complex web of norms, assumptions, meanings, practices, and social dynamics that make up culture and context. Their development is both influenced by culture (e.g., drawing on data reflecting perspectives from 21st century internet culture, while having been developed by engineers operating in a particular milieu) and in turn influences culture (e.g., shaping the kind of content people create and consume, while integrating into an ever-expanding set of production pipelines across society).

While this position is increasingly accepted as orthodoxy within the field of AI, it can rely on a limited definition of culture. In practice culture is often treated as a secondary consideration, like a coat of paint or dash of seasoning that modifies the more ``fundamental'' aspects of the model: for example, as a bias to debug \cite{bolukbasi2016man,tao2024cultural}, a constraint for generalizing from one context to another \cite{cao2023assessing}, a parameter in an ethical dilemma \cite{de2021driverless}, or a source of variability in user preferences \cite{ge2024culture}. These approaches operationalize culture as a variable to be measured---often implying that it is an optional parameter to include in model evaluation, rather than a foundational aspect of the model's functioning. 

However, the most prominent frontier models---many of which not only power popular general-purpose AI systems (e.g., ChatGPT, Gemini) but also underlie task-specific applications (e.g., Microsoft Copilot, Notion)---are not specialized systems designed, and often marketed, to solve targeted, well-defined tasks.  They are marketed as general systems built to generate a variety of cultural artifacts in a vast space of possible contexts. Cultural considerations are inextricable both from how these models are developed and from the open-ended, dialogic interfaces in which they are used. It is therefore crucial that we ask: How can we most effectively evaluate GenAI as a cultural technology?

In this Perspective, we offer an account of culture informed by the humanities. We argue that evaluation methods in AI often overlook an important conception of culture: not as a variable to be measured, but as a dynamic, contested space where social meaning is made \cite{geertz1973interpretation, hall1997representation, klein2025provocations}. This way of looking at culture challenges a core assumption in standard practices for AI benchmarking---that model performance is best understood through universal, standardized tasks with convergent solutions or goals \cite{raji2021ai}. While this approach works for well-defined tasks where ``success'' can be codified into a single, unique interpretation, culture is not this kind of task. 

To illustrate the challenge of evaluating cultural outputs, consider the act of writing a letter, painting a portrait, composing a song, cooking a meal, penning a journal entry, or even talking with a friend. While it is possible to assign a quantitative score to these outputs to describe how well the task was performed, that approach can miss the point of these activities in crucial ways. For example, reducing cultural activities to a set of proxy variables can trivialize them \cite{zhou2025culture}, while scalable ``thin'' metrics are often insufficient to capture key aspects of what makes them meaningful \cite{kommers2025meaning}. The structure of these tasks is such that the primary question is not about assessing how closely they cleave to a canonical ground truth. Rather it is about arbitrating among multiple, possibly conflicting, interpretations of their meaning within a specific frame of reference. This requires us to think about culture as an intrinsically different kind of ``task'' from those by which a model's performance has traditionally been judged.

Our position is that, as AI systems are increasingly deployed to (co-)produce cultural outputs, it is imperative that our methods of evaluation reflect the interpretive dimensions needed to characterize them more fully. To address this, we introduce hermeneutics---a core tradition in the humanities concerned with the theory and practice of interpretation---as a theoretical foundation for understanding and evaluating GenAI systems \cite{mohr2015toward, romele2020digital, rebera2025hidden}. Having grappled with these questions for decades, if not centuries, the conceptual infrastructure of the humanities (via hermeneutics) can help articulate the grounds on which a given interpretation can be considered legitimate. Providing such an account of the interpretive nature of GenAI systems is a crucial step towards improving the way we design and evaluate them.

Thus, we present computational hermeneutics as an emerging framework offering an interpretive approach to the evaluation of GenAI systems. We argue that GenAI can, and should, be understood as ``doing'' interpretation in ways that reflect the entanglement of culture in their input, processes, and outputs. We use this active phrasing---to ``do'' interpretation---to reflect the fact that interpretive considerations are inextricably bound up in the processes and structures underlying these systems. The data, architecture, and algorithms of these systems are not static mirrors reflecting back invariant, disinterested projections of the world. They must be understood as comprising interpretive stakes, decisions, and processes. On the other hand, it is nonetheless crucial to recognize that GenAI do not ``do'' interpretation in the same way as humans and to avoid unduly imbuing them with anthropomorphic intentions \cite{devrio2025taxonomy, akbulut2024all}.

With this in mind, we offer three hermeneutic challenges that are inherent to such interpretive processes: situatedness, plurality, and ambiguity. Each of these already exists in one form or another in contemporary AI \cite{sorensen2024position, akbulut2025century, lazar2023ai}. We further this existing work by suggesting how these challenges can be brought together within a hermeneutic frame. Finally, we offer three principles for developing hermeneutic methods of evaluating GenAI: that benchmarks should be iterative, not one-off; include people, not just machines; and measure cultural context, not just model output.

\section{Computational Hermeneutics}

Interpretation is the methodological bedrock of the humanities \cite{geertz1973interpretation}. Generally speaking, what humanists do when studying cultural artifacts---whether a novel, historical event, or painting---is to construct an interpretation: an analysis of that artifact's meaning within its social or historical context. But this approach comes with an inherent challenge. How do we know whether a given interpretation is a good one? Hermeneutics is the method, justification, or separate interpretive process which gives credence or legitimacy to the original interpretation \cite{caputo2018hermeneutics, ricoeur1981hermeneutics}. This concept is foundational across many disciplines and practices, from legal and literary studies \cite{szondi1995introduction,levinson1988interpreting} to debates in philosophy and aesthetics \cite{simpson2020hermeneutics, rosen2003hermeneutics}. It arises, in one form or another, whenever scholars confront epistemological problems of meaning. 
%Where one scholar encounters a text, they will produce an interpretive account of the text's meaning; where many scholars encounter a text, they will debate the interpretive process by which they arrived at such an account.

A core concept within this tradition is the ``hermeneutic circle'' \cite{schleiermacher1998schleiermacher, dilthey1989introduction, heidegger1927being, gadamer1960truth}. This describes the interpretation of an artifact as an iterative process between understanding the meaning of a specific part of the artifact and the meaning of the artifact as a whole. For example, one could iteratively analyze the imagery depicted in a given line or stanza of a poem, then update one's conception about what the poem means in general---each time using the updated general theory to analyze the specific line, and vice versa. While the term is varied in its usage, what it typically means to analyze something hermeneutically is to engage in (and provide an account of) this iterative process of interpretation.

As applied to contemporary AI, we offer a notion of computational hermeneutics in two senses. The first sense is that AI models are fundamentally interpretive in a way that makes hermeneutic problems unavoidable; these challenges are intrinsic to GenAI's flexible production of sophisticated cultural artifacts such as texts and images. To categorize their outputs as binary ``right'' or ``wrong'' responses presents a similar profile of problems as asking whether \textit{Anna Karenina} is a superior novel to \textit{Jane Eyre}, whether the spiritual life prescribed in Laozi's \textit{Tao Te Ching} is the right one, or whether Andy Warhol's soup can paintings were a critique, rather than a celebration, of American consumerism. Judgments on these matters are possible, but they depend crucially on the underlying assumptions of one's interpretive processes.

The second sense is that interpretive evaluation requires us to look at both specific and general aspects of the models, in the tradition of the hermeneutic circle. These models have both a general architecture (e.g., pre-training, vector representations, fine-tuning), as well as specific dialogic interactions with human users (e.g., context windows, prompts). We must look at both system-level and context-specific generalizations in interpreting the outputs of these models. Roughly speaking, partial analysis maps onto the ``Chat'' in ChatGPT, while holistic analysis maps onto the ``GPT.'' Though these separate parts are interrelated, it is crucial to draw distinctions required for the evaluation of each on their own terms \cite{dobson2019critical, ringler2024computation}.

\subsection{Hermeneutic Challenges for AI}

With this framing in mind, we present three hermeneutic challenges for GenAI: situatedness, plurality, and ambiguity. Each of these challenges take aspects of a model that may seem arbitrary, peripheral, or in need of optimization---and re-centers those apparently accidental features as significant choices worthy of theoretical reflection. We take addressing these challenges to be the main difference between accounting for culture as a variable versus culture as a site of social meaning-making.

\subsubsection{Situatedness: Meaning only emerges in context.}

A core principle across many (if not all) of the humanities is that context is key. What this expresses, typically, is that to interpret the meaning of a cultural artifact, one must look at the historical or social context in which it has been made, used, or perceived \cite{gadamer1960truth}. For example, a contemporary reader of \textit{Huckleberry Finn} will inevitably have a different relation to the text from a reader in the 19th century America of the book’s original publication. When the frame of reference shifts, so does the meaning. Cultural products are always generated within the bounds of a particular historical, cultural, or communicative context. This is the ``situatedness'' of meaning: an interpretation always takes a particular point of view, even if that perspective is only stated implicitly. 

It can be easy to overlook this in contemporary AI interfaces, which often present the model as speaking from a god’s eye point of view---that of the disembodied model which has seen, read, and synthesized more information than any one human ever could \cite{Hemment2025doing}. No such epistemically totalitarian ``view from nowhere'' exists in any legitimate sense \cite{haraway1988situated}. Within a hermeneutic frame, the point is not to build and evaluate models that aim to achieve this universal, monolithic perspective. Rather it is for the specific perspective being offered to be clearly identified and understood as just that: a specific perspective. For example, recent work has empirically demonstrated how GenAI systems can collapse perspectives into an idealized form, showing how further mechanisms are needed to maintain the individuation of distinct perspectives \cite{heuser2025cultural}.

\subsubsection{Plurality: One person's bias is another person's values.}

Interpretation is inherently plural, because different communities rely on distinct frameworks for making sense of the world. What appears as meaningful artistic expression to one group may seem inappropriate or offensive to another; what counts as authoritative fact in one tradition may be dismissed as unsubstantiated assertion in another. As is widely held in the humanities, multiple valid interpretations can coexist without requiring resolution into a single ``correct'' reading. Any AI model intended for use in different cultural contexts must grapple with the observation that what looks like arbitrary cultural bias from one perspective is often the same thing that gives a sense of meaning and value in another.

AI systems face this challenge directly because they serve users with distinct values while being trained on materials whose authors often disagree. Generative models are therefore both one and many: reflecting specific curatorial decisions, but also containing contradictory voices \cite{desai2024archival, sharma2024facilitating, veselovsky2025hindsight}. Recent work on pluralistic, thick, or full-stack alignment recognizes that human values naturally conflict and advocates for systems that can accommodate this diversity \cite{sorensen2024position, lazar2023ai, lowe2025full}. However, while pluralistic alignment focuses on adjusting model behavior to reflect different values, the deeper challenge lies in how we evaluate such systems. Standard evaluation frameworks assume convergent solutions---that there is a standard candle against which model performance can be definitively compared. Cultural tasks, by contrast, do not converge to single solutions: success cannot be determined by proximity to a ground truth but must account for the legitimacy of multiple interpretations within their respective contexts. This requires fundamentally rethinking evaluation from measuring accuracy to assessing appropriateness across different cultural frameworks \cite{leibo2024theory}. For example, what is viewed as AI ``slop'' in one context may be valued as a legitimate source of meaning or seen as having aesthetic resonance in another \cite{kommers2025slop}.

\subsubsection{Ambiguity: Interpretations naturally conflict.}

In hermeneutics, meaning is not something that exists as a fixed property of a text or cultural artifact, inertly awaiting discovery. Rather, meaning emerges through what Gadamer calls the ``fusion of horizons''---the dynamic interaction between the interpreter's background and the artifact being interpreted \cite{gadamer1960truth}. This process is intrinsically ambiguous. The space of possible mappings between potentially relevant features of the interpreter's background and the artifact is combinatorially large, and therefore a definitive interpretation is not computationally tractable. To offer a particular kind of interpretation (e.g., feminist, post-colonial, techno-optimist) is to ease this intractability by specifying an a priori constraint on which features to consider. More generally, Gadamer emphasizes the role of ``play'' in interpretation---that creative, open-ended consideration of tensions between different meanings offers a way of exploring this space of interpretive possibilities. It is therefore crucial that ambiguity be maintained in articulating this interpretive space, rather than being flattened into a specific mode of interpretation.

Ambiguity has long been of interest in AI, often with the goal of resolving it \cite{navigli2009word}. Semantic disambiguation tasks, for instance, aim to determine which meaning of a polysemous word is intended in a given context---clarifying whether ``light'' is used to signify illumination or weight. Such tasks are crucial for many applications, but they represent only one way of engaging with ambiguity. When AI systems generate cultural outputs---whether composing poetry, engaging in dialogue, or creating visual art---the goal is not necessarily to eliminate semantic uncertainty but to work productively within it \cite{gaver2003ambiguity}. A poem that resolves all its ambiguities loses much of its interpretive richness; a conversation that admits only one reading of each utterance becomes sterile \cite{empson1930seven}. However, current evaluation frameworks often treat this ambiguity as noise to be minimized rather than a generative resource \cite{yadav2021comprehensive}. While semantic disambiguation tasks can be useful, elimination of ambiguity is not the only---or even the primary---goal when it comes to cultural outputs. Instead, evaluation should assess how well systems navigate ambiguity productively, maintaining the interpretive flexibility that enables meaningful cultural engagement across diverse contexts \cite{veselovsky2025localized, leibo2024theory}. For example, recent approaches have sought to tease out the inherently multiplicitous perspectives contained within GenAI systems, developing processes for negotiating among conflicting viewpoints held within the model architecture \cite{li2024can}.

\section{Generative AI systems as ``Context Machines''}

In this section, we argue that GenAI systems ``do'' interpretation as a fundamental capacity \cite{dobson2022vector}---and therefore evaluation of their performance is subject to the three hermeneutic challenges described above.Even so, it is important to note that the interpretive processes underlying these systems are distinct from those of human interpreters \cite{placani2024anthropomorphism}; while conversational systems may superficially adopt the voice of a human perspective, they should not be mistaken as inveterately human \cite{peter2025benefits}. For instance, such interpretive processes take place both internally within a model, as well as dialogically in their interactions with people. Providing a more comprehensive account of the interpretive nature of these systems is a crucial step towards improving the way we design and evaluate them.

We posit that GenAI systems can be broadly understood as ``context machines.'' At core, GenAI systems are designed to answer the question: given the current context, what is the next relevant token, pixel, or other value? This ability to consolidate a broader set of contextual cues into a unified representation is supported by a variety of architectural features---but most notably by vector space embeddings \cite{kozlowski2019geometry, stoltz2021cultural, ethayarajh2019contextual}. Such embeddings are a means of encoding highly sophisticated co-occurrence statistics \cite{turney2010frequency}. In language models, they are learned by poring over vast corpora of text \cite{mikolov2013distributed, pennington2014glove}. In vision models, vectors of pixel values are often encoded as feature maps capturing edges, textures, and semantic patterns \cite{bengio2012unsupervised, mihai2021learning}. Decoding these embeddings is also an interpretive act. This process is often probabilistic, accommodating a plurality of possible interpretations \cite{james2013introduction, yang2023diffusion}. Informally, these vectors are designed to capture the ``meaning'' of words or images; more concretely, they are a highly nuanced way of describing the context in which a word is likely to occur. 

Generative models work as well as they do because (as is a common refrain in the humanities) context matters---so much so that if you get it right, a lot of other important things follow. Vector space embeddings are therefore subject to a similar question as humanistic inquiry: How do we know whether a given interpretation, as encoded by an embedding, is a good one? Accordingly, GenAI systems are faced with the three hermeneutic challenges described above: the outputs of these systems are situated (the ``meaning'' of one token is defined relationally within the context of other tokens); plural (there are multiple legitimate interpretations of what counts as the next most likely token); and ambiguous (the probabilistic decoding process maintains rather than resolves semantic uncertainty). 

Our position is that Generative AI systems both ``do'' interpretation, and that they can do it better. For example, the self-attention mechanism of the transformer architecture can be read as a way of relating partial and holistic interpretations \cite{vaswani2017attention}. It allows the model to iteratively update its understanding of individual tokens based on their relationship to the broader sequence, and vice versa---in other words, the hermeneutic circle in action.

\subsection{AI systems don't just ``read in'' context; they help create it.}

Interpretation does not only occur in isolation within GenAI models; these systems also co-construct interpretations in collaboration with humans \cite{frauenberger2019entanglement}. A hermeneutic perspective on AI is not just about building systems that can interpret like humans, as a substitute or proxy for human expertise. Rather it is about recognizing how interpretation itself emerges through interaction between humans and machines. In this view, interpretive capacity arises not only within the model but through the design of interactions and interfaces that frame it.

The effects of this collaboration are bidirectional. From human to machine, people decide what data the systems are trained on \cite{desai2024archival}; formulate objective functions that reflect a specific set of goals, values, and assumptions \cite{lazar2023ai}; fine-tune system behavior through mechanisms like reinforcement learning from human feedback \cite{ouyang2022training}; and ``engineer'' prompts in order to elicit certain kinds of responses \cite{chen2025unleashing}. At multiple layers of the system, human annotators---who can themselves offer conflicting interpretations \cite{frenda2024perspectivist}---can provide feedback on ambiguous cases, rank responses, or supply preference scores, effectively staging a dialogue where the AI's provisional interpretations can be contested and refined. 

From machine to human, AI systems affect important mental capacities like metacognition \cite{tankelevitch2024metacognitive}; elicit different assumptions about relational norms (e.g., AI as assistant vs therapist \cite{earp2025relational}); act as thought-partners, for example by summarizing documents people would otherwise have to read---or skim---in full \cite{collins2024building}; shape human responses by explaining their own decisions \cite{doshi2017towards}; and enable novel kinds of experience, such as certain creative practices \cite{hemment2024experiential, caramiaux2022explorers, murray2021emergent}. Examples of how this approach has been employed include applications of assemblage thinking to study how AI is deployed \cite{tseng2023assemblage}, as well as design frameworks that incorporate interpretive practices into multiple steps of the development process \cite{andres2025scenario}. Together, humans and GenAI systems form an interpretive feedback loop. Far from a separate isolated entity that the system merely ``reads in,'' AI systems can exert a direct influence on the cultural context in which they operate.

\section{Operationalizing Hermeneutics in AI}

Typically, AI benchmarking assumes universal, standardized tasks with convergent solutions \cite{raji2021ai,chang2024survey, eriksson2025can}---an approach fundamentally at odds with a hermeneutic perspective on culture. While benchmarks are key drivers of progress in AI, they often do not offer especially strong standards for what they purport to measure \cite{kapoor2024ai, mcintosh2025inadequacies, reuel2024betterbench, schlangen2021targeting}. Furthermore, the implicit goal of benchmarking is often not to develop stronger metrics for specialized cases (though see \cite{chiu2024computational, underwood2025can}) but something more like one-task-suite-to-rule-them-all, a comprehensive assessment that would give an unequivocal, decisive answer to the question of which model is better at what \cite{raji2021ai, koch2024protoscience, alzahrani2024benchmarks, ethayarajh2020utility, srivastava2023beyond}. 

Our hermeneutic framing challenges this paradigm by reimagining the kinds of questions that can be asked with AI benchmarks: shifting from standardized questions about accuracy to contextual ones about meaning. From this perspective, no such comprehensive task suite can be developed, because the ``task'' of creating cultural outputs means too many different things in too many different contexts. Attempts to standardize cultural production into a comprehensive assessment often seek to scrub away this context; we advocate that such context must be embraced. We offer three ways of making AI benchmarks that better reflect a hermeneutic lens on culture---by making them iterative, not just one-off; including people, not just machines; and measuring cultural context, not just model output.

\subsection{Benchmarks should be iterative, not just one-off.}

The hermeneutic circle suggests that interpretation depends on an iterative process between part and whole. By contrast, benchmarks typically apply a score---often scalar values such as accuracy, precision, recall, F1, or BLEU scores \cite{chang2024survey, eriksson2025can}---to quantify the model's performance in a given domain. Hermeneutics benchmarking suggests two modifications that can be made to this approach. 

First, evaluation is both limited and unreliable when it scores performance based on a single prompt \cite{mizrahi2024state}. By contrast, cultural outputs are always part of an evolving conversation, whether a literal dialogue or as a part of a broader evolutionary process \cite{brinkmann2023machine}. Evaluation should accordingly be iterative, unfolding over multiple prompts or exchanges that reflect the evolving interpretive context. 

Second, evaluation must take into account both the model as whole and the specific dialogic frame in which a given output is elicited. For example, the focus of benchmarking on aggregate metrics indicating average performance rather than instance-by-instance evaluations limits generalizability \cite{burnell2023rethink}. Overall, hermeneutic evaluations should seek to iteratively assess both the model's holistic capabilities, as well as its behavior in specific circumstances.

Existing benchmarks have begun to incorporate multi-turn iterative processes into their evaluation practices. Notable examples include assessments of chatbot capabilities in more than a dozen distinct tasks which evaluate performance over the course of an interaction with a human interlocutor \cite{bai2024mt}, as well as assessments of perceived anthropomorphism in language models which show that the interpretation of model behavior as social (e.g., ``relationship building'' via empathic, emotionally-validating responses) only take place after multiple turns of interaction \cite{ibrahim2025multi}.

\subsection{Benchmarks should include people, not just machines.}

The interpretive processes underlying GenAI are inextricably bound up in collaboration with the people using them \cite{messeri2024artificial}. Benchmarks should therefore not just consider AI performance in isolation but ought to also measure the effects of different interactive configurations. For example, current approaches to the assessment of creativity in narrative generation range from automated metrics to expert human judgment \cite{boisson2025automatic, marco2025reader, chakrabarty2024art}; but these often treat creativity as a model property rather than a relational phenomenon. 

A hermeneutic approach would evaluate how human-AI collaboration produces  interpretations, examining not just outputs but the interpretive dialogue that generates them. This builds on a wide range of efforts in AI evaluation which increasingly recognize that benchmarks cannot be divorced from their communicative context \cite{chiu2024computational, denton2020bringing, weidinger2023sociotechnical, weidinger2024star}. Overall, hermeneutic evaluation requires benchmarks that assess interactivity rather than isolated performance, examining not just outputs but the interpretive dialogue that generates them.

This practice is increasingly adopted in AI benchmarking. For example, assessments of harms from GenAI systems are sensitive to a larger range of potential issues---such as social manipulation or cognitive over-reliance---only by evaluating the model capabilities in conjunction with their use by a human \cite{ibrahim2025towards}. Likewise, a recent benchmark looking at cultural expectation incorporates over 10,000 human annotations, reflecting norms and judgments based on the lived experience of people from a given cultural domain \cite{nayak2025culturalframes}.

\subsection{Benchmarks should measure cultural context, not just model output.}

Individual interpretations of meaning depend on cultural context \cite{kommers2025sense}---yet standard evaluation practices treat context as secondary to model performance metrics. Thin signals of like/dislike, positive/negative, or use/disuse cannot provide this contextual grounding \cite{kommers2025meaning}. Rather, we need hermeneutic approaches for putting contextual use cases on equal footing with general model capacities.

Partially, this is simply a suggestion to evaluate AI in the context in which it will be used \cite{liao2023rethinking, messeri2024artificial, tomaszewska2024position, malaviya2025contextualized, akbulut2025century}. For example, frameworks like HELM recognize the need for contextually dependent approaches beyond accuracy \cite{liang2022holistic}. This can help address issues with current benchmarks, such as failure to capture real-world utility \cite{ott2022mapping}, or by adapting general processes to better fit situational needs \cite{staufer2025audit}.

But more pointedly, digging deeper into contextualized scenarios allows us to probe different aspects of the model. Rather than asking whether a response is correct, hermeneutic evaluation can assess how and why a response achieves appropriateness within its specific cultural framework \cite{leibo2024theory, bhutani2024seegull}. Evaluation must treat cultural context not as a constraint on model performance, but as the medium through which such performance emerges.

Some benchmarks are beginning to incorporate these kinds of contextual markers. For instance, a recent benchmark contrasts socio-cultural norms for Chinese vs American viewers of AI-generated videos \cite{varimalla2025videonorms}. Similarly, a recent dataset organizes feedback from human raters based on demographic information, allowing for distinction in cultural judgments \cite{rastogi2025whose}. In summary, it is worth noting that the strongest of exemplars of hermeneutic evaluation tend to adopt all three recommendations: they are iterative, incorporate human participants, and sensitive to sociocultural variation.

\section{Discussion}

Computational hermeneutics represents a potential shift in how we conceptualize GenAI systems. Rather than treating culture as a variable to be controlled or optimized away, we propose recognizing it as a foundational aspect of how these systems operate. This reframing transforms GenAI from answer-generating machines into interpretive partners---systems designed to engage with the situatedness, plurality, and ambiguity that characterize individual and collective human meaning-making.

In this article, we have focused on the evaluation of GenAI systems via benchmarks. We offer this as a potentially effective means by which scholars with a humanistic background can help shape the direction of technical development in AI. Benchmarks are an important part of how the field of AI progresses and understands its own progress. However, in practice benchmarks often fall short of meaningfully assessing what they purport to measure \cite{kapoor2024ai, mcintosh2025inadequacies, reuel2024betterbench, schlangen2021targeting}, and it is widely acknowledged that better benchmarks are needed to support ethical and effective development of AI \cite{blagec2023benchmark, ren2024safetywashing, zhao2025assessing}. One possible systemic cause of this is proxy failure \cite{john2024dead}: that the field’s monocultural over-reliance on standardized performance metrics is inadequate to capture the kinds of things we really want AI to do \cite{koch2024protoscience, kommers2025meaning, zhou2025culture}. This gives researchers with expertise in operationalizing tricky social or cultural concepts a useful lever for influencing this technology’s metrics for success. But while we have focused on evaluation, this is not the only way to employ a hermeneutic perspective in AI. For example, on-going debates look at the cultural and social underpinnings of a model’s training data \cite{mihalcea2025ai, ravichander2025information}.

More generally, the hermeneutic tradition points to how powerful technological systems cannot be considered only in isolation, without appreciation of their environmental and societal consequences; this is a juncture at which crucial debates are being held and to which we hope a hermeneutic perspective can contribute. We offer the emerging framework of computational hermeneutics as a potential means of rethinking how we evaluate AI from the ground up---as a set of technologies that does not just participate in culture by accident, but as systems which fundamentally shape, and are shaped by, cultural meaning. 

\section{Declarations}

\subsection{Author Contributions}

C.K. proposed the initial idea and led the collaborative drafting of the manuscript. D.H. oversaw and contributed to the process in a senior role. All authors contributed to the development of the manuscript. 

%\subsection{Competing Interests}

%We declare that the authors have no competing interests as defined by Nature Portfolio, or other interests that might be perceived to influence the results and/or discussion reported in this paper.

\subsection{Acknowledgments}

This study was funded by the Arts and Humanities Research Council UK and Lloyd's Register Foundation. These funders played no role in the writing of this manuscript. 

%%%%%%%%%%%%%%%%%%%%%%%%%%%%%%%%%%%%%%%%%%%%%%%%%%%%%%%%%%

\newpage

\small{
\bibliography{Main}
}

\end{document}